\documentclass[lettersize,journal]{IEEEtran}
\usepackage{amsmath,amsfonts}
\usepackage{algorithmic}
\usepackage{algorithm}
\usepackage{array}
\usepackage[caption=false,font=normalsize,labelfont=sf,textfont=sf]{subfig}
\usepackage{textcomp}
\usepackage{stfloats}
\usepackage{url}
\usepackage{verbatim}
\usepackage{graphicx}
\usepackage{pifont}
\usepackage{booktabs}
\usepackage{multirow}
\usepackage{cite}
\begin{document}

\title{A Weak Signal Learning Dataset and Its Baseline Method}

\author{Xianqi~Liu, Xiangru~Li, Lefeng~He, and Ziyu~Fang%
\thanks{Xianqi~Liu is with the School of Artificial Intelligence, South China Normal University, Foshan 528225, People's Republic of China 
(e-mail: 2023025175@m.scnu.edu.cn).}%
\thanks{Xiangru~Li, Lefeng~He, and Ziyu~Fang are with the School of Computer Science, South China Normal University, Guangzhou 510631, People's Republic of China
(e-mail: xiangru.li@gmail.com; 17607644285@163.com; fangziyushiwo@126.com).}%
}


\maketitle

\begin{abstract}
Weak signal learning (WSL) is a common challenge in many fields like fault diagnosis, medical imaging, and autonomous driving, where critical information is often masked by noise and interference, making feature identification difficult.  Even in tasks with abundant strong signals, the key to improving model performance often lies in effectively extracting weak signals.  However, the lack of dedicated datasets has long constrained research.  To address this, we construct the first specialized dataset for weak signal feature learning, containing 13,158 spectral samples.  It features low SNR dominance (over 55\% samples with SNR below  50) and extreme class imbalance (class ratio up to 29:1), providing a challenging benchmark for classification and regression in weak signal scenarios.  We also propose a dual-view representation (vector + time-frequency map) and a PDVFN model tailored to low SNR, distribution skew, and dual imbalance.  PDVFN extracts local sequential features and global frequency-domain structures in parallel, following principles of local enhancement, sequential modeling, noise suppression, multi-scale capture, frequency extraction, and global perception.  This multi-source complementarity enhances representation for low-SNR and imbalanced data, offering a novel solution for WSL tasks like astronomical spectroscopy.  Experiments show our method achieves higher accuracy and robustness in handling weak signals, high noise, and extreme class imbalance, especially in low SNR and imbalanced scenarios.  This study provides a dedicated dataset, a baseline model, and establishes a foundation for future WSL research.
\end{abstract}


\section{Introduction}
\IEEEPARstart{S}{upervised} machine learning aims to uncover the mapping relationships and intrinsic patterns between input features and output labels.  As one of the core tasks in artificial intelligence, it has achieved remarkable progress in model architectures and learning algorithms, and has been widely applied across numerous scientific and engineering domains \cite{choudhary2017comprehensive}. Examples include fault diagnosis, medical image analysis, geological exploration, natural language processing, image restoration, and autonomous driving. However, although available methods perform well on the majority of samples, the key challenges in both research and practice often lie in those few samples that are difficult to correctly identify or predict. 

Although these difficult samples are limited in quantity, they often critically determine the overall system's reliability, safety, and practical value. Such samples are typically characterized by weak features, extreme scarcity, or discriminative features that occupy a very small portion of the input. They may be referred to as ``Critical Few Samples" or ``Edge Cases". For instance, in medical imaging, rare or atypical early-stage malignant tumor samples may lead to serious clinical consequences (missing the optimal treatment window) if misdiagnosed \cite{naeem2024breast}; In autonomous driving, pedestrians or non-standard obstacles suddenly appearing under extreme weather conditions can easily cause accidents if misidentified \cite{zhang2023idt}. These cases not only reveal the boundaries of model generalization capabilities but also highlight the urgent need for highly robust and sensitive learning methods. This class of problems is categorized in this paper as the ``weak signal learning“ (WLS) problem, and we conduct exploration around specilized dataset construction and benchmark methods for it.

However, current research on weak signal machine learning mostly depends on conventional datasets, where only a small number of weak signal samples are manually selected for exploration. This approach has obvious limitations: weak signal samples are severely insufficient in terms of quantity, diversity, and problem coverage. These limitations make them difficult to systematically support in-depth studies on model generalization and robustness. It is worth noting that some studies have begun to touch upon certain aspects of WSL. For instance, datasets constructed for small object segmentation and tracking tasks\cite{sang2023small,zhu2024tiny} focus on scenarios where discriminative features occupy an extremely low proportion in the samples. Meanwhile, research on imbalanced learning and long-tailed distributions\cite{wang2024more,yang2021delving,LI2024110701} addresses challenges arising from sample scarcity and uneven class distribution. Although these works involve certain characteristics of weak signals (such as faint features or sample scarcity) to varying degrees, they do not systematically cover the multiple manifestations of weak signals (e.g., low signal-to-noise ratio, feature masking, dynamic interference) and lack a unified and standardized benchmark evaluation environment. Therefore, constructing a dedicated dataset that encompasses various types of weak signals with significant challenges and diversity is particularly important. This paper aims to fill this gap.

To address this gap, we constructed a WSL Dataset (WSLD) for regression estimation and classification tasks. This dataset exhibits the following three distinctive characteristics:
\begin{enumerate}
    \item \textbf{Low Signal-to-Noise Ratio Sample Dominance}: Over 55\% of the samples in WSLD have a signal-to-noise ratio (SNR) below 50 (Fig. \ref{fig:mae_snrg_distribution}), with an even higher proportion of low-SNR samples in minority classes. Samples of this type inherently exhibit weak and poorly discriminative features. These characteristics make them difficult to model accurately. Yet, they often have a decisive impact on the overall system performance — analogous to the missed detection of atypical early-stage tumors in medical imaging or the misidentification of sudden obstacles under extreme weather conditions in autonomous driving, where erroneous judgments can lead to serious consequences. Therefore, WSLD effectively serves to test the boundaries of a model's capability in weak signal perception and generalization performance.
    \item \textbf{Dual Imbalance in Regression and Classification Tasks}:  The WSLD exhibits severe imbalance in both regression and classification configurations. In regression tasks, the distribution of samples is highly non-uniform in the parameter spaces of $T_{\texttt{eff}}-\log~g$ and [Fe/H]-[C/Fe] (Fig. \ref{fig:distribution_density_prediction_accuracy}). To our knowledge, no dedicated imbalanced dataset specifically designed for regression tasks currently exists in the literature. For classification tasks, there is a substantial disparity in class sample sizes, with the majority class containing 6,640 samples and the minority class comprising only 229 samples. This high degree of imbalance tends to cause models to be biased toward the majority class, leading to a significant decline in recognition accuracy for minority classes. However, these minority-class samples often correspond to critical objects in scientific discovery (such as unknown celestial types) or rare cases in medical diagnosis, making their correct identification highly valuable. The simultaneous coverage of both regression and classification imbalance constitutes an important distinctive feature of the WSLD.
    \item \textbf{High-Dimensional Features and Large-Scale Samples}: The dataset comprises 13,158 high-dimensional samples, each with 3,450 features, significantly exceeding the typical scale of imbalanced datasets provided on the UCI platform\cite{uci_repo}—where most datasets range from only 4 to 166 features. This high-dimensional feature structure not only more realistically reflects the complexity of information sources in practical applications but also provides models with a richer feature space to uncover underlying patterns. It is particularly suitable for large-scale modeling and fine-grained representation requirements in weak-signal scenarios.
\end{enumerate}

The dual challenges of low signal-to-noise ratio (SNR) and extreme class imbalance form the core difficulties in regression and classification tasks within WSL. These two factors intertwine and mutually reinforce each other, significantly increase the complexity of model training. In high-noise scenarios, substantial variations in signal strength render inherently weak discriminative features easily overwhelmed by background noise, making it difficult for models to capture reliable key features or parameter correlation patterns. Meanwhile, extreme class imbalance further induces structural bias in models, causes them to over-rely on the distribution characteristics of majority class samples during training, thereby severely compromises their ability to represent weak signals and recognition sensitivity in minority classes\cite{mayabadi2022two}.  More critically, these two issues often coexist in practical applications: low SNR samples are predominantly concentrated in minority classes. This coupled effect not only widens the model's generalization gap but also imposes higher demands on the robustness and adaptability of existing learning algorithms.

To facilitate further exploratory research, this paper proposes a dual-view data representation scheme and a Parallel Dual-View Fusion Network (PDVFN) as a baseline processing solution. One view consists of the sample's corresponding vector, while the other view is a time-frequency image derived via short-time Fourier transform of the vector. The complementarity of this dual-view representation is profoundly reflected in its synergistic exploration of the intrinsic information structure of astronomical spectral signals. The vector representation carries the original physical information of the signal, with its sequential structure directly corresponding to the physical relationship between wavelength and flux. Therefore, this representation accurately preserves the position, profile, and local morphology of spectral lines. The time-frequency representation, on the other hand, reveals structural information of the spectral signal in the frequency domain. The PDVFN model extracts local sequence features and global frequency-domain structures through parallel processing of ACR and PMTF modules, makes the final decision through feature splicing and fusion. This architecture leverages multi-source information complementarity to enhance the representation capability of low signal-to-noise ratio and imbalanced distribution data, provides novel solutions for WSL tasks such as astronomical spectroscopy. The ACR module achieves context-aware, attention-focused, and robust sequential representation extraction through local feature enhancement, sequential dependency modeling, and noise suppression. The PMTF module accomplishes scale invariance, long-range dependency modeling, and time-frequency structure identification via multi-scale structure capture, frequency-domain feature extraction, and global pattern perception.

In summary, the contributions of this work can be summarized as follows:
\begin{enumerate}
    \item Constructed a dedicated dataset for WSL, named WSLD. This dataset comprises 13,158 samples and is characterized by two primary challenging features: low SNR predominance and extreme class imbalance. It provides a unified and reliable benchmark evaluation platform for regression and classification tasks in weak signal scenarios.
    \item A dual-view representation (vector + time-frequency feature map) and the PDVFN model are proposed. This design leverages multi-source information complementarity to enhance the model's representational capacity for low signal-to-noise ratio and imbalanced distribution data, provides a novel solution for WSL tasks such as astronomical spectra.
    \item The effectiveness of the proposed method has been systematically validated across multiple experimental settings. It demonstrates significant superiority over existing methods. This provides a strong benchmark framework for WSL tasks.
\end{enumerate}

\section{RELATED WORKS}

WSL widely exists in scientific and engineering fields. Its core challenge lies in the fact that critical information often appears with extremely low intensities and can easily be masked by relatively high-amplitude background noise or complex interferences. These characteristics make feature extraction and pattern recognition exceptionally difficult. This issue is particularly prominent in a series of important applications, including fault diagnosis \cite{wang2023incipient,liu2025bearing,ding2023deep,yu2025anc,liu2021wind}, medical image analysis \cite{jiang20213d,wang2025mrsenet}, seismic signal processing \cite{wei2022hybrid,liu2023consecutively,zhong2024shbgan,iqbal2023deepseg,zhong2023rmchn,zhang2024dual,tian2022novel}, natural language processing \cite{lan2020redundant,leem2024selective}, image restoration \cite{fan2023lacn}, and motion estimation in satellite videos \cite{wang2023multi}, etc. The effective learning of weak signals is not only a critical bottleneck for improving performance in various tasks but also a vital pathway for uncovering intrinsic data patterns and promoting reliable model generalization in complex environments. Consequently, WSL has recently become a research focus in numerous application fields and is increasingly gaining theoretical attention in the machine learning and pattern recognition communities.

WSL involves two interconnected yet distinct elements: the inherently weak but valuable signals and the interfering noise. According to their respective focus, existing methods can be broadly categorized into weak signal perception–oriented (WSPO) and noise suppression–oriented (NSO) approaches. WSPO methods aim to capture subtle features within complex backgrounds by optimizing feature extraction or incorporating attention mechanisms to enhance sensitivity to minute pattern variations. In contrast, NSO methods improve weak signal discernibility by reducing irrelevant noise components through noise-robust representations or targeted filtering techniques. Although differing in focus, both share the same goal: improving models’ ability to exploit weak signals in high-noise environments. In practical applications such as fault diagnosis and medical image analysis, these two types of methods are often combined to enhance intelligent system performance in complex scenarios.

In the researches on weak signal perception, several innovative solutions have been proposed in recent years from various perspectives. For example, Lan et al. \cite{lan2020redundant} introduced a channel attention mechanism into a convolutional encoder-decoder network for adaptively screening feature responses. This approach significantly enhances the model's sensitivity to weak signal by capturing local time-frequency details of speech signals amid complex background noise.  Wang et al. \cite{wang2023incipient} addressed the challenge of weak and easily obscured early fault signals by proposing an unsupervised feature fusion method based on a Deep Extreme Learning Machine Denoising Autoencoder. This method effectively extracts weak fault features in multi-scale, multi-channel environments and improves signal discriminability. To tackle the difficulty of estimating weak target motion from low SNR satellite videos, Wang et al. \cite{wang2023multi} constructed a multi-frame sparse self-learning PWC network. By incorporating motion consistency constraints and a sparse self-learning strategy, the network enhances the capability of capturing and tracking weakly moving objects. In the field of fault diagnosis, Liu et al. \cite{liu2025bearing} proposed an adaptive decomposition algorithm based on frequency marginal spectrum local maxima. This method effectively increased the model's sensitivity to weak stoppage features by isolating weak stoppage components and selecting the most information-rich modes. These studies advance the development of weak signal perception capabilities from multiple perspectives—including feature extraction, multimodal fusion, and structural modeling, act as critical technical pathways in this field.

In the field of weak signal enhancement, researchers have proposed various methods in recent years. These investigations significantly improve the discernibility and usability of weak signals across different tasks. For example,  Wei et al. \cite{wei2022hybrid} introduced a hybrid loss-guided coarse-to-fine model for enhancing the reconstruction quality of weak signals in seismic data. This model first reconstructs dominant signal components via a coarse network and then employs a refinement network with weighted masking to focus on recovering weak signal regions. Leem et al. \cite{leem2024selective} proposed a selective enhancement strategy for speech emotion recognition. This strategy targets noise-corrupted speech by enhancing only the weak acoustic features that significantly contribute to emotion recognition while preserving features with strong noise resistance. It is shown that this strategy can maintains system robustness while enhancing weak signals. Liu et al. \cite{liu2023consecutively} addressed the challenge of weak signal recovery in cases of consecutive missing seismic data by proposing a reconstruction method based on wavelet transform and Swin Residual Network. By integrating multi-scale wavelet decomposition and local-global modeling capabilities into a U-Net, this method significantly improves the reconstruction accuracy of weak signals. Zhong et al. \cite{zhong2024shbgan} developed a Super-Resolution Hybrid Bilateral Attention Generative Adversarial Network. This network enhances the detailed reconstruction of faint geological structures in seismic images by incorporating a hybrid bilateral attention mechanism and dilated convolutions. Ding et al. \cite{ding2023deep} proposed a deep time-frequency learning-based enhancement method for weak fault diagnosis in rotating machinery. This method utilizes physically meaningful time-frequency masks to achieve precise enhancement of fault resonance bands. It is shownt that this method improves diagnostic performance while enhancing model interpretability. Jiang et al. \cite{jiang20213d} addressed the issue of low SNR in 3D neuronal microscopy images by proposing a method combining a ray-shooting model with a Dual-Channel Bidirectional Long Short-Term Memory (DC-BLSTM) network. The DC-BLSTM effectively enhances weak neuronal structural signals while suppressing background noise by transforming the 3D segmentation task into a sequence learning problem. These studies systematically advance the development of weak signal processing technology from multiple perspectives, including reconstruction, enhancement, and selective optimization. They not only deepen the understanding of weak signal enhancement mechanisms but also provide important methodological support for the utilization of faint information in complex environments.

In WSL, noise suppression constitutes another pivotal research direction that aims to enhance the relative prominence of weak signals, improve their discernibility and usability by suppressing background interference. This category of methods has achieved remarkable progress across multiple domains. Iqbal et al. \cite{iqbal2023deepseg} proposed the DeepSeg framework for seismic signal processing by jointly optimizing sparse signal representation and noise suppression through time-frequency modeling. It is shown that this scheme can effectively extracts faint seismic events in highly noisy environments. For rotating machinery fault diagnosis, Yu et al. \cite{yu2025anc} developed ANC-Net for enhancing fault recognition robustness across varying noise levels and operating conditions. ANC-Net is a multi-scale active noise cancellation network based on discrete wavelet transform. This method leverages residual learning to extract domain-invariant features. In biomedical signal analysis, Wang et al. \cite{wang2025mrsenet} introduced MrSeNet for improving the detectability of subtle abnormal features. The MrSeNet is a multi-resolution deep network with attention mechanisms. This scheme suppresses noise interference in electrocardiogram signals through multi-scale feature fusion and channel attention recalibration. To address noise amplification in low-light image enhancement, Fan et al. \cite{fan2023lacn} designed LACN, a lightweight attention-guided ConvNeXt network that incorporates parameter-free attention modules and multi-stage feature fusion to effectively mitigate noise accumulation during the brightening process. For wind turbine blade bearing fault diagnosis, Liu et al. \cite{liu2021wind} proposed a Bayesian-augmented Lagrangian (BAL) algorithm, which reformulates the filtering problem as a Bayesian sub-optimization process to successfully extract faint fault signals against strong noise backgrounds. To improve weak signal recovery in complex noisy environments, Tian et al. \cite{tian2022novel} developed a parallel attention-guided multi-branch residual network that integrates multi-scale convolution and attention mechanisms to suppress multiple types of noise while recovering consecutive weak signals. In distributed acoustic sensing (DAS) data denoising tasks, Zhong et al. \cite{zhong2023rmchn} proposed a residual modularized cascaded heterogeneous network that combines long- and short-path feature learning strategies to effectively enhance background noise suppression and deep weak signal recovery. For detecting deep weak signal from DAS vertical seismic profiling (VSP) data, Zhang et \cite{zhang2024dual} further introduced a dual-attention denoising network that significantly improves deep weak signal recovery accuracy through multi-scale feature extraction and feature reweighting. These approaches have systematically advanced noise suppression techniques from diverse perspectives, including sparse representation, residual learning, attention mechanisms, multi-scale modeling, and Bayesian optimization. These investivations  provide solid supports for weak signal detection and recovery in complex environments.

Despite the considerable progress achieved by available researches in feature enhancement, noise suppression, and model architecture design, they still exhibit evident limitations in complex scenarios involving intense noise interference, extremely low SNR, or severe sample imbalance. Specifically, current models fall short in terms of the granularity of weak feature extraction, coherent modeling of long-range temporal dependencies, and effective utilization of global contextual information, thereby failing to fully meet the elevated requirements for robustness and generalization in WSL tasks.

\section{A Weak Signal Data Set, its Characteristics and Challenges}
\label{sec:DataSet}

The objective of this section is to construct a weak-signal dataset (WSLD) and use it as a framework to systematically integrate the key pathway from defining ``weak signals" to designing learning models. We begin by addressing the question ``What constitutes a weak signal?" (Section~\ref{sec:DataSet_definitionOfWeakSignal}), establishing the logical starting point for the study. We then present the procedures for constructing and processing a high-quality WSLD (Sections~\ref{sec:DataSet_construction} and \ref{sec:DataSet_process}), providing a reproducible data benchmark. Through a multi-dimensional analysis of the dataset (Section~\ref{sec:DataSet_statisticsAndAnalysis}), we reveal its intrinsic characteristics and patterns. Finally, these data-driven insights are translated into core challenges and guiding principles for model design (Section~\ref{sec:DataSet_modelDesignChallenges}), offering clear directions and fundamental foundations for the development of next-generation WSL algorithms.

\subsection{Definition of Weak Signals}
\label{sec:DataSet_definitionOfWeakSignal}

In numerous artificial intelligence applications—such as astrophysics, medical diagnosis, industrial fault monitoring, and financial risk control—researchers commonly face a fundamental challenge: how to identify and extract subtle yet decision-critical information patterns from complex and noise-contaminated data. These patterns are generally referred to as ``weak signals". Although their importance has been widely recognized, the academic community has yet to establish a unified and universally accepted definition of weak signals, as their specific connotations often vary depending on the research domain and problem context.

\subsubsection{The Connotation of Weak Signals from a Multidisciplinary Perspective}

By reviewing how different disciplines interpret weak signals, we can summarize their commonalities and distinctive characteristics, thereby lay a foundation for constructing a broadly applicable and representative dataset.

\begin{itemize}
    \item In the fields of traditional signal processing and sensors, weak signals are most intuitively manifested as physical signals with low SNR. In such cases, the effective information is severely interfered with or masked by strong background noise, making detection and extraction extremely challenging\cite{wang2013current}.
    \item In the fields of industrial and financial forecasts, the notion of weak signals shifts from purely physical properties to the early emergence and predictive value of patterns. It refers to subtle yet highly informative early indications or statistical deviations that arise in system operations or market dynamics. Examples include slight anomalous variations in vibration spectra that indicate potential equipment faults or minor fluctuations in financial time series that foreshadow significant trend reversals\cite{berrouche2024local, thavaneswaran2022deep}. Although these signals are inconspicuous, they are crucial for achieving accurate early warning and forward-looking decision-making.
    \item In the fields of business and sociological analysis, weak signals are more abstract and are often regarded as early signs of emerging trends or systemic transformations. They are scattered across massive volumes of text, social media, or user behavior data, appear as low-intensity, small-scale, and often unstructured information fragments that may foretell future shifts in consumer preferences or social hot topics\cite{jun2017visualization, silva2024unsupervised}.
\end{itemize}

\subsubsection{Definition, Core Challenges, and Dataset Construction Principles of Weak Signals in the Context of Machine Learning}

Based on the considerations of multidisciplinary perspectives outlined above, we can formulate the following operational definition for weak signals within the context of machine learning tasks: Weak signals refer to informational patterns present in the input data that exhibit weak discriminative power and are easily obscured by dominant features or noise, yet exert important or even decisive influence on the model's final decisions or outputs. In real-world machine learning applications, effective learning of weak signals primarily confronts two interconnected core challenges, which constitute the defining characteristics of weak-signal problems:

\begin{itemize}
    \item The weakened representataion and obscuration on feature level. The effective features of weak signals occupy an extremely small proportion of the overall feature space, even often appear vague, indistinct, or lacking clear structures. These phenomena may arise from low signal-to-noise ratios in the physical world, limitations in observation conditions, or the intrinsic properties of the features themselves. As a result, conventional feature extraction mechanisms (e.g., standard convolution kernels or global attention mechanisms) struggle to reliably and accurately localize and model these faint discriminative cues within high-noise or highly correlated contexts.
    \item The imbalance and sparsity on data level. Samples carrying weak signals (such as critical minority categories or sparse regions representing specific phenomena in parameter space) typically constitute a very small portion of the overall dataset. These samples often lie near decision boundaries or in the long-tail regions of the data distribution. Consequently, during standard training, the optimization objectives (e.g., empirical risk minimization) naturally bias the model toward dominant, high-frequency patterns in the majority data. The bias leads to the neglect of weak signal samples. This imbalance and sparsity not only impair the model’s ability to perceive and learn weak signals but also cause degraded generalization and biased decision-making in rare yet critical scenarios.
\end{itemize}

Therefore, to systematically investigate weak-signal learning and construct a broadly representative dataset capable of effectively supporting algorithmic exploration, it is essential to comprehensively address the dual challenges at both the feature and data levels. At the feature level, the weak-signal problem is fundamentally characterized by a weakened representation of discriminative features, most directly quantified by a low signal-to-noise ratio (SNR)—when key feature signals are faint and information-sparse, or when background noise and interference components are complex and intense, the SNR significantly decreases, forming a typical weak-signal scenario. At the data level, the challenges manifest as imbalanced distributions and sparsity, specifically including uneven sample distributions in the parameter space and quantitative imbalances among classes. These two aspects collectively constitute the core issues to be resolved in constructing a weak-signal dataset.
\begin{itemize}
    \item Nonuniform distribution in parameter space. In regression or density estimation problems, the number of samples in certain specific regions of the estimated parameters is substantially lower than in other regions. This leads to a notable decline in the model's predictive performance within these ``sparse regions," primarily because the model lacks sufficient samples to learn the data dynamics characteristic of such areas.
    \item Class imbalance. In classification tasks, some classes have far fewer samples than others. The phenomenon results in a coexistence of ``majority" and ``minority" classes. This imbalance skews the optimization process toward the majority classes, drastically reduces the model’s recognition performance for minority classes—i.e., its ability to capture weak signals.
\end{itemize}

Based on these considerations, a high-quality WSL dataset should systematically cover the scenarios with various levels of SNR, differing degrees of data distribution unevenness, and multiple levels of class imbalance. Such a dataset would provide a comprehensive and solid benchmark for evaluating and advancing WSL algorithms.

\subsection{Construction of Weak-Signal Dataset}
\label{sec:DataSet_construction}

As previously discussed, a comprehensively representative weak-signal dataset should systematically encompass weakened and obscured feature expressions, as well as imbalanced and sparse data distributions. To this end, this work selects the search for Carbon-Enhanced Metal-Poor (CEMP) stars in large-scale spectroscopic surveys as the application context for constructing the dataset \cite{Fang_2025}.

CEMP stars are a class of rare celestial  objects with extremely low metallicity (e.g., iron) but abnormally high carbon abundance relative to metallicity. They are considered as important ``fossil relics" for understanding the chemical enrichment of the early universe and the nature of the first-generation stars. Systematic searches for such stars holds great astrophysical significance. This task embodies the two core challenges of WSL:
\begin{itemize}
    \item At the feature level, challenges arise from weakened feature and feature obscuration. Large-scale surveys typically limit single-exposure time to improve observation efficiency. This procedure results in spectra with low SNRs. The key discriminative features of CEMP stars—molecular absorption lines associated with carbon (e.g., the G-band)—appear weak and indistinct in such spectra, easily drowned out by noise. However, these massive low-SNR datasets also offer new opportunities to discover rare stellar objects that traditional high-precision surveys cannot easily capture, while simultaneously imposing higher demands on information extraction techniques.
    \item At the data level, there exist issues of distribution imbalance and sparsity. CEMP stars are extremely rare, typically comprising less than 1\% in all stars, act as a typical case of extreme class imbalance. Moreover, these stars are unevenly distributed within the multidimensional parameter space. They often appear in narrow regions and exhibit strong data sparsity.
\end{itemize}

Currently, three main technical approaches are commonly used to identify CEMP stars from large spectroscopic datasets: model-fitting-based parameter estimation, two-stage filtering and end-to-end classification \cite{Fang_2025}.

However, in weak signal scenarios, all of the above-mentioned approaches face severe challenges. The first two methods heavily depend on the reliability of parameter estimation on low-SNR samples, while parameter estimation itself constitutes a typical weak-signal regression problem prone to error propagation. The end-to-end classification methods, on the other hand, suffer from the extreme class imbalance and data sparsity, which can cause models to overlook rare classes. Therefore, constructing a benchmark dataset specifically designed for this problem is crucial for systematically evaluating the robustness, accuracy, and generalization ability of different methods in weak signal scenarios.

The WSLD dataset is constructed using the LAMOST (Large Sky Area Multi-Object Fiber Spectroscopic Telescope)\cite{Zhao_2012, cui2012} DR11 v1.1 low-resolution spectral library, with sample labels derived from multiple authoritative sources\cite{ApJS:Abdurro:2022, Aoki2022ApJ, LiHaiNing2022ApJ, PASJ:Suda:2008, MNRAS:Suda:2011, MNRAS:Suda:2013, PASJ:Suda:2017, yuan2020dynamical}. During dataset construction, we employed the data cross-matching tool TOPCAT\cite{taylor2005topcat} to efficiently integrate multi-source data, ensuring label consistency and reliability. Meanwhile, a rule-driven labeling strategy was adopted to assign class labels to each sample, maintain interpretability and consistency of classification criteria. Specifically, the stellar parameters of each spectrum—effective temperature ($T_\texttt{eff}$), surface gravity ($\log g$), metallicity ([Fe/H]), and carbon abundance ([C/H])—were collected from published catalogs and literature. Based on these parameters, we calculated the carbon-to-iron ratio [C/Fe] = [C/H] - [Fe/H] and assigned class labels according to the metallicity threshold for metal-poor stars ([Fe/H] $<$ -1.0) from \cite{ARAA:Beers:2005} and the CEMP criteria proposed in \cite{ApJ:Aoki:2007}.

In practice, we first determine whether a target star is metal-poor based on its [Fe/H] value. If it is metal-poor, we further assess whether it meets the CEMP criteria proposed in \cite{ApJ:Aoki:2007}; if so, it is classified as CEMP, otherwise as carbon-normal metal-poor (CnMP). For [Fe/H] $>$ -1.0, the sample is assigned to the non-metal-poor (NMP) category. The three categories are numerically labeled as NMP = 0, CEMP = 1, and CnMP = 2. This rule-driven labeling method, grounded in physical definitions and literature standards, ensures that sample categories in WSLD are consistent, interpretable, and scientifically valid, thereby provides a reliable foundation for model training and performance evaluation.

To construct a benchmark dataset that accurately reflects the challenges of WSL, we not only ensured the intrinsic data quality but also designed a partitioning strategy that supports effective learning and fair evaluation in complex scenarios. Specifically, the WSLD dataset was randomly divided into training ($\mathcal{D}{\text{train}}$, 9,210 samples), validation ($\mathcal{D}{\text{val}}$, 1,316 samples), and test ($\mathcal{D}_{\text{test}}$, 2,632 samples) subsets in a 7:1:2 ratio. The core purpose of random partitioning is to ensure that the intrinsic noise patterns, quality fluctuations, and intra-class feature distributions are evenly represented across all subsets. This prevents distributional bias introduced by artificial splitting and guarantees that the ``weak signal–noise" conditions encountered during training remain consistent with those in testing, thereby laying a foundation for robust learning across diverse feature distributions. However, given the extreme class imbalance (CEMP stars account for <1\%), simple random partitioning could lead to insufficient or missing rare-class samples in certain subsets. To mitigate this, we maintained relative class balance across training, validation, and test sets, ensuring adequate representation of rare classes (CEMP stars) in each subset. This treatment allows the validation and test sets to reliably assess model generalization in the presence of truly sparse weak signals. In summary, through a combination of random partitioning and class distribution control, the WSLD dataset provides a benchmark that faithfully reflects real-world uncertainty while ensuring fair evaluation—enabling systematic assessment of algorithmic robustness and generalization under weak-signal, high-noise, and extremely imbalanced conditions.

\subsection{Data Preprocessing}
\label{sec:DataSet_process}

To ensure that the original observational data are suitable for machine learning tasks in weak-signal scenarios and to minimize the interference of non-astrophysical factors on model learning, we performed a series of rigorous preprocessing operations on the raw LAMOST spectra. These steps aim to eliminate the effects of instrumental responses, observing conditions, and noise. These preprocessings bring all spectra onto a consistent physical baseline and numerical scale to highlight their intrinsic astrophysical characteristics. The followings provide a detailed description of each preprocessing step, including its methodological design, underlying rationale, and contributions to improving data quality and model performance.

\begin{enumerate}
    \item Wavelength correction. Using the radial velocity estimates provided by LAMOST, each spectrum is corrected to the rest frame according to the following equation:
    \begin{equation}
    \lambda' = \frac{\lambda}{1 + \frac{RV}{c}}
    \end{equation}
    where $\lambda'$ is the corrected wavelength, $\lambda$ is the original wavelength, $RV$ is the radial velocity of the corresponding spectrum, and $c$ is the speed of light. Because of the Doppler effect caused by the motion of the Earth and the intrinsic radial motion of stars, the observed spectral wavelength grids exhibit systematic shifts. The primary purpose of this step is to unify all spectra in the rest-frame coordinate system. By doing this preprocessing, the variations in line positions introduced by relative motion of the sources is eliminated. This correction serves as the physical foundation for subsequent alignment, comparison, and feature matching across spectra. Without it, identical absorption lines would appear at different wavelengths in different spectra. This phenomenon can introduce noise and significantly impair the model’s ability to recognize and learn stable spectral features.
    \item Linear interpolation resampling. Within the largest common wavelength interval, linear interpolation is performed in logarithmic wavelength space using a fixed step size, standardizes each spectrum to 3,450 data points. The wavelength grids of raw spectra may differ slightly between observational batches or celestial targets. The primary objective of resampling is to achieve strict numerical alignment across all spectra, ensure that each pixel input to the machine learning model corresponds to exactly the same physical wavelength. Interpolation in logarithmic space is chosen because spectral features such as absorption line widths are more stable on a logarithmic wavelength scale. This operation ensures data format consistency and provides the regularized grid structure required for convolutional neural networks and other grid-based models to operate effectively.
    \item Denoising (median filtering). A median filter with a window size of three pixels is applied to smooth the spectral curves.Low SNR is one of the main challenges in weak-signal datasets, where bad pixels and random impulse noise can obscure subtle astrophysical features. Median filtering, as a nonlinear filtering technique, effectively removes sharp impulse noise (e.g., bad pixels caused by cosmic ray hits on CCDs) while preserving the edge morphology of absorption lines. This avoids the blurring of spectral features that may occur with linear filters such as Gaussian smoothing. As an initial denoising step, it provides a cleaner data foundation for subsequent continuum normalization.
    \item Continuum normalization. A continuum is a smooth and continuous dependencies of light intensity on wavelength. The purpose of continuum normalization is to highlight the features of absorption and emission lines and lay the foundation for machine learning by eliminating observational effects such as atmospheric extinction and instrument response. This study employs a fifth-order polynomial to fit the continuum. Continuum normalization is performed by dividing the original spectrum by this fitted continuum, ultimately yielding a normalized spectrum with a continuum background of approximately 1.
    \item Secondary denoising and standardization. Flux outliers are removed based on the $3\sigma$ rule, followed by Z-score standardization.After continuum normalization, local non-impulsive noise and residual flux fluctuations may still remain. To address these problems, for each normalized spectrum \(\textbf{X} = \{x_1, \dots, x_L\}\), the mean flux \(\mu\) and standard deviation \(\sigma\) are computed as:
    \begin{equation}
    \mu = \frac{1}{L} \sum_{i=1}^L x_i
    \end{equation}
    \begin{equation}
    \sigma = \sqrt{\frac{1}{L} \sum_{i=1}^L (x_i - \mu)^2}
    \end{equation}
    Then, each flux less than \(\mu - 3\sigma\) or greater than \(\mu + 3\sigma\) is replaced by the mean value \(\mu\). Finally, standardization is performed by converting each spectral flux \(x_i\) as following:
    \begin{equation}
    s_i = \frac{x_i - \mu}{\sigma}, \quad i = 1, \dots, L
    \end{equation}
    This step performs a fine-grained secondary cleaning by statistically identifying and correcting residual outliers after normalization. Therefore, this procedure further smooths the data. The subsequent Z-score normalization (subtracting the mean and dividing by the standard deviation) adjusts all spectra to have zero mean and unit variance. This operation aligns with the default assumptions of many machine learning optimizers, accelerates model convergence and prevents unstable gradients caused by large value disparities across dimensions. Therefore, this procedure improves both numerical stability and training efficiency.
\end{enumerate}

In summary, the proposed systematic preprocessing pipeline—covering wavelength correction to establish a consistent physical reference, linear resampling for data alignment, multi-stage noise suppression, and flux normalization—substantially enhances the consistency, purity, and astrophysical expressiveness of the spectral data. This pipeline not only provides a reliable and well-structured foundation for weak-signal feature learning but also bridges the methodological gap between astrophysical data processing principles and the practical requirements of machine learning models.

\subsection{Investigations on the characteristics of WSLD dataset}
\label{sec:DataSet_statisticsAndAnalysis}

In weak-signal learning tasks, the intrinsic characteristics of a dataset fundamentally determine a model’s ability to capture subtle yet essential patterns, while also exposing potential learning bottlenecks and generalization risks. This section systematically analyzes the structural properties of the WSLD dataset from three key perspectives: the distribution of signal-to-noise ratios, data imbalance, and the complexity of high-dimensional feature structures. Through this analysis, we aim to uncover the major challenges inherently posed by the data itself and provide a foundation for subsequent model design and optimization. These challenges include the modeling difficulties associated with low-SNR samples, underfitting in extremely sparse classes and parameter regions, and the representational complexity arising in high-dimensional spaces. Our analysis reveals that the WSLD dataset represents a typical weak-signal learning scenario, characterized by low-SNR skew, pronounced imbalance across classes and parameter spaces, and intricate high-dimensional feature relationships. While these characteristics exacerbate the difficulty of model learning, they also offer clear directions and unique value for advancing algorithmic innovation under weak-signal conditions. The following analysis of specific dataset characteristics further elucidates the origins and implications of these challenges. 
\begin{enumerate}
    \item SNR Distribution: Dominance of Low-SNR Samples and Their Systematic Impact on Model Errors. The WSLD dataset exhibits a pronounced bias toward low-SNR. As illustrated in Fig. \ref{fig:mae_snrg_distribution}(b), more than 55\% of the samples have an SNR below 50, forming the majority of the dataset. This distribution poses a fundamental challenge to model learning: as shown in Fig. \ref{fig:mae_snrg_distribution}(a), all methods exhibit significantly higher parameter estimation errors in low-SNR regions than in high-SNR regions, with estimation accuracy consistently improving as SNR increases. Classification performance is similarly affected by SNR; under low-SNR conditions, the discriminative accuracy across all categories generally declines, with minority classes—whose intrinsic discriminability is already weak—being most severely impacted. The complexity of this issue arises from the substantial overlap between low-SNR samples and critical minority classes. Such samples possess weak features and low discriminability, making them highly susceptible to being overlooked or misclassified during training, which in turn introduces evaluation bias. In real-world applications, this bias can lead to serious consequences, such as missed detection of early lesions in medical imaging or misrecognition of extreme scenarios in autonomous driving. Therefore, the interplay between low SNR and class structure constitutes a primary challenge in the WSLD dataset. 
    \item Data Imbalance: Dual Imbalance in Parameter and Class Spaces, and Its Learning Challenges. The WSLD dataset exhibits a pronounced data imbalance in both regression and classification tasks. This imbalance is characterized by uneven distributions in both parameter and class spaces. In regression tasks, the distribution of samples across key physical parameter spaces is highly non-uniform, as demonstrated in both the effective temperature-surface gravity ($T_{\text{eff}}-\log~g$) and chemical abundance ([Fe/H]-[C/Fe]) spaces. As shown in Fig. \ref{fig:distribution_density_prediction_accuracy}(a), in the chemical abundance space, samples cluster in areas of medium-to-high metallicity and low carbon abundance, whereas those in high carbon abundance regions are exceedingly rare. Therefore, this dataset forms a typical imbalanced distribution. A noteworthy phenomenon is that the model's parameter estimation error increases in regions with sparse samples (Fig. \ref{fig:distribution_density_prediction_accuracy}(b)), demonstrating that data scarcity directly impacts learning performance. This density-dependent performance pattern is also observed in classification tasks, where recognition accuracy similarly decreases in data-sparse regions. Additionally, in classification tasks, the impact of class imbalance is particularly significant. The three classes contain 6,640 (NMP), 6,290 (CnMP), and 229 (CEMP) samples, respectively. Especially, the CEMP stars accounting for less than 2\%, form a typical long-tailed distribution. Although the numbers of CnMP and NMP samples are comparable, the extreme scarcity of CEMP samples results in poor recognition performance for this class. It is worth emphasizing that, despite their rarity, CEMP stars carry critical astrophysical information, and the ability to accurately identify them serves as a key indicator of a model's potential for scientific discovery.
    \item High-Dimensional Feature Structure: Representational Potential and Computational Challenges in Weak-Signal Environments. The WSLD dataset provides a crucial foundation for weak-signal learning in terms of both scale and dimensionality. With 13,159 samples, each represented by 3,450 features, it significantly surpasses typical imbalanced datasets from the UCI repository in feature dimensionality—where most datasets range from 4 to 166 features, with only a few exceptions like Toxicity (1,203 features). 
    This high-dimensional feature structure, on the one hand, reflects the inherent complexity of information sources in real-world problems and offers a broad feature space for models to uncover latent patterns under weak-signal conditions. On the other hand, it imposes greater demands on feature selection, regularization design, and computational efficiency. In such high-dimensional spaces, effectively capturing subtle yet discriminative signals while suppressing noise interference constitutes a core challenge in weak-signal learning.      
\end{enumerate}

Through the analysis of the three core characteristics of the WSLD dataset, we systematically reveal the structural properties of data in weak-signal learning scenarios and their fundamental implications for machine learning models. The analysis demonstrates that low-SNR skew, multi-level imbalance, and high-dimensional features collectively form a challenge framework of substantial practical relevance: low SNR weakens the model’s discriminative foundation, data imbalance distorts its learning objectives, and high-dimensionality further amplifies the complexity of representation and computation. While these characteristics increase the modeling difficulty, they also delineate clear research directions—developing representations robust to low SNR, designing loss functions and sampling mechanisms adapted to imbalanced structures, and constructing efficient models capable of handling high-dimensional features. The analysis of this dataset not only provides empirical evidence for understanding the intrinsic challenges of weak-signal learning but also establishes a clear technical roadmap for subsequent model development and algorithmic innovation.

\subsection{Model Design Challenges}
\label{sec:DataSet_modelDesignChallenges}

The WSLD dataset holds substantial research significance in weak-signal learning. Its distinctive characteristics—including a skewed low-SNR distribution, dual imbalance across both parameter and class spaces, and a high-dimensional feature structure (as analyzed in Section~\ref{sec:DataSet_statisticsAndAnalysis})—provide a complex and realistic experimental environment for exploring regression, classification, and weak-signal learning tasks. These properties not only impose considerable challenges on model design but also present crucial opportunities for developing robust models capable of handling low SNR, sparse classes, and high-dimensional features. Consequently, the WSLD dataset contributes to advancing frontier technologies that rely on weak-signal modeling, such as astronomical spectral analysis, medical image processing, and autonomous driving.

\begin{figure}[!t]
    \centering
    \setlength{\belowcaptionskip}{-85mm} \captionsetup[subfigure]{skip=2pt, font=scriptsize} 
    \subfloat[{\scriptsize $T_\texttt{eff}$ MAE}\label{fig:teff_mae}]{\includegraphics[width=0.48\linewidth]{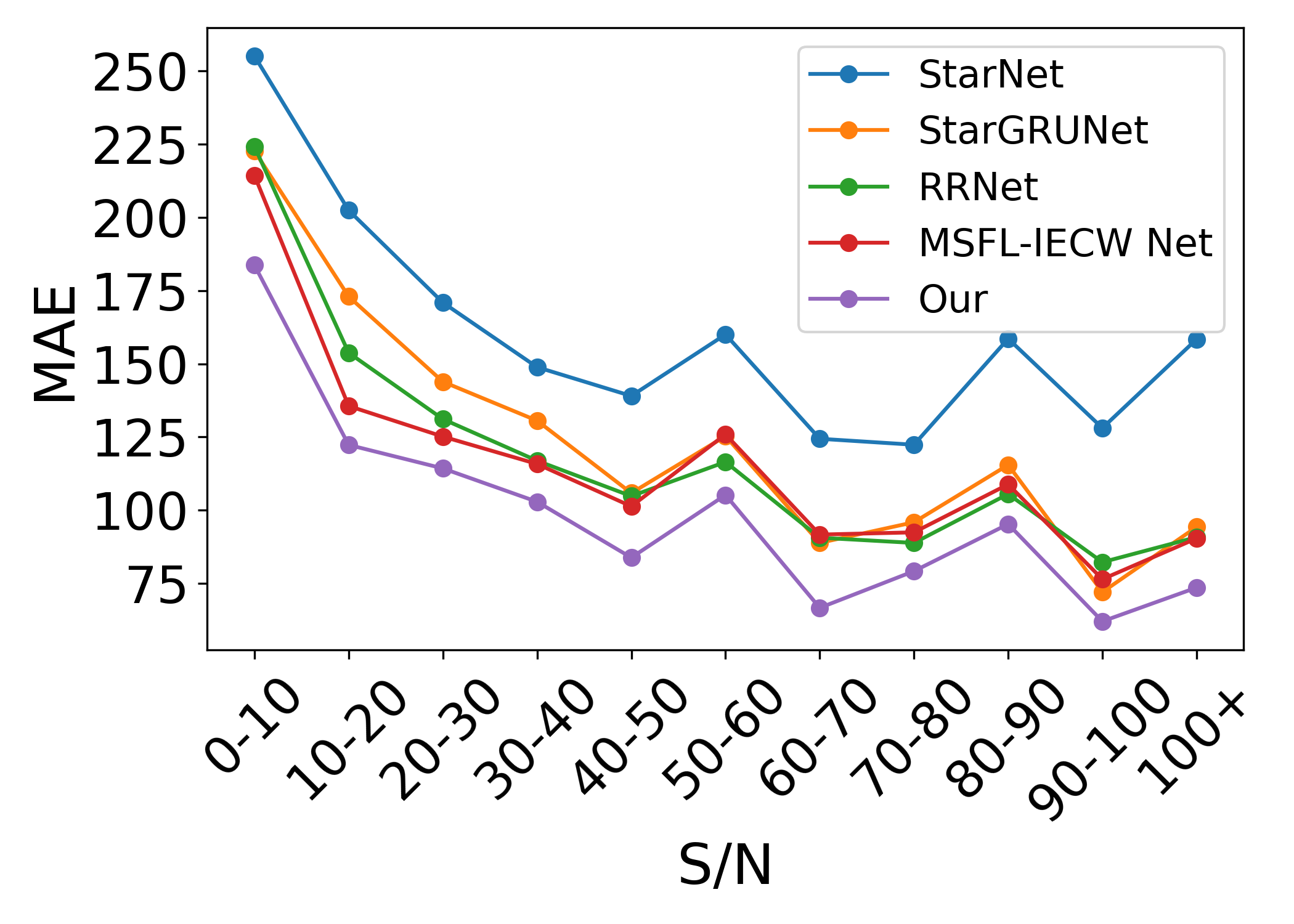}}%
    \hfil
    \subfloat[{\scriptsize SNR distribution}\label{fig:snr_dist}]{\includegraphics[width=0.48\linewidth]{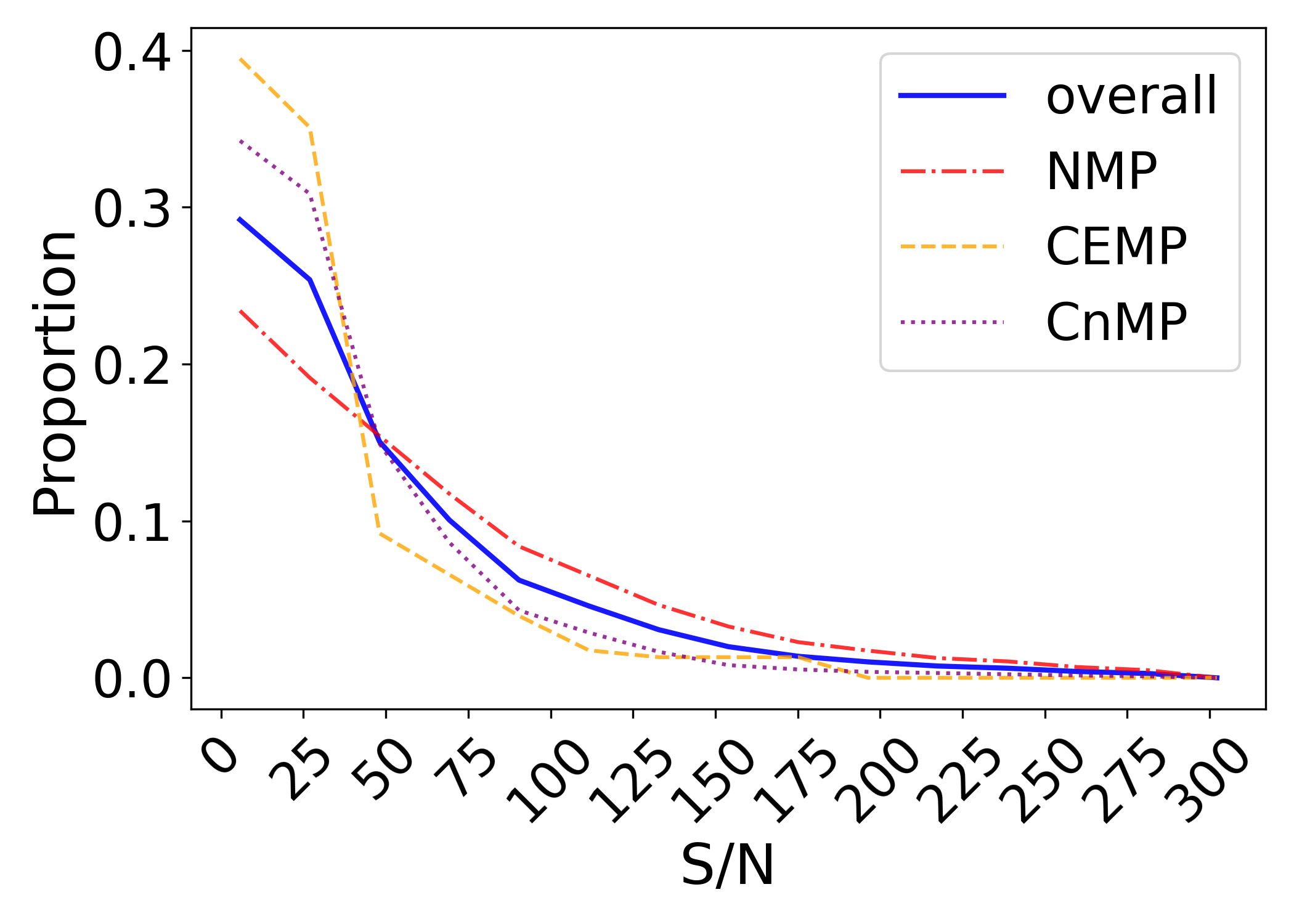}}

    \caption{The dependencies of learning performance on data quality SNR and the distribution of the data quality. The estimation performance decreases dramatically with the decrease of SNR, and the appearence probability of data increase with the decrease of SNR. The performance patterns of $\log~g$, [Fe/H], [C/H] and classification are similar with the $T_\texttt{eff}$.}
    \label{fig:mae_snrg_distribution}
\end{figure}

\begin{figure}[!t]
    \centering
    \setlength{\belowcaptionskip}{-55mm} 
    \captionsetup[subfigure]{skip=2pt, font=scriptsize} 
    \subfloat[{\scriptsize \texttt{[Fe/H]}-\texttt{[C/Fe]} density}\label{fig:feh_cfe_density}]{\includegraphics[width=0.48\linewidth]{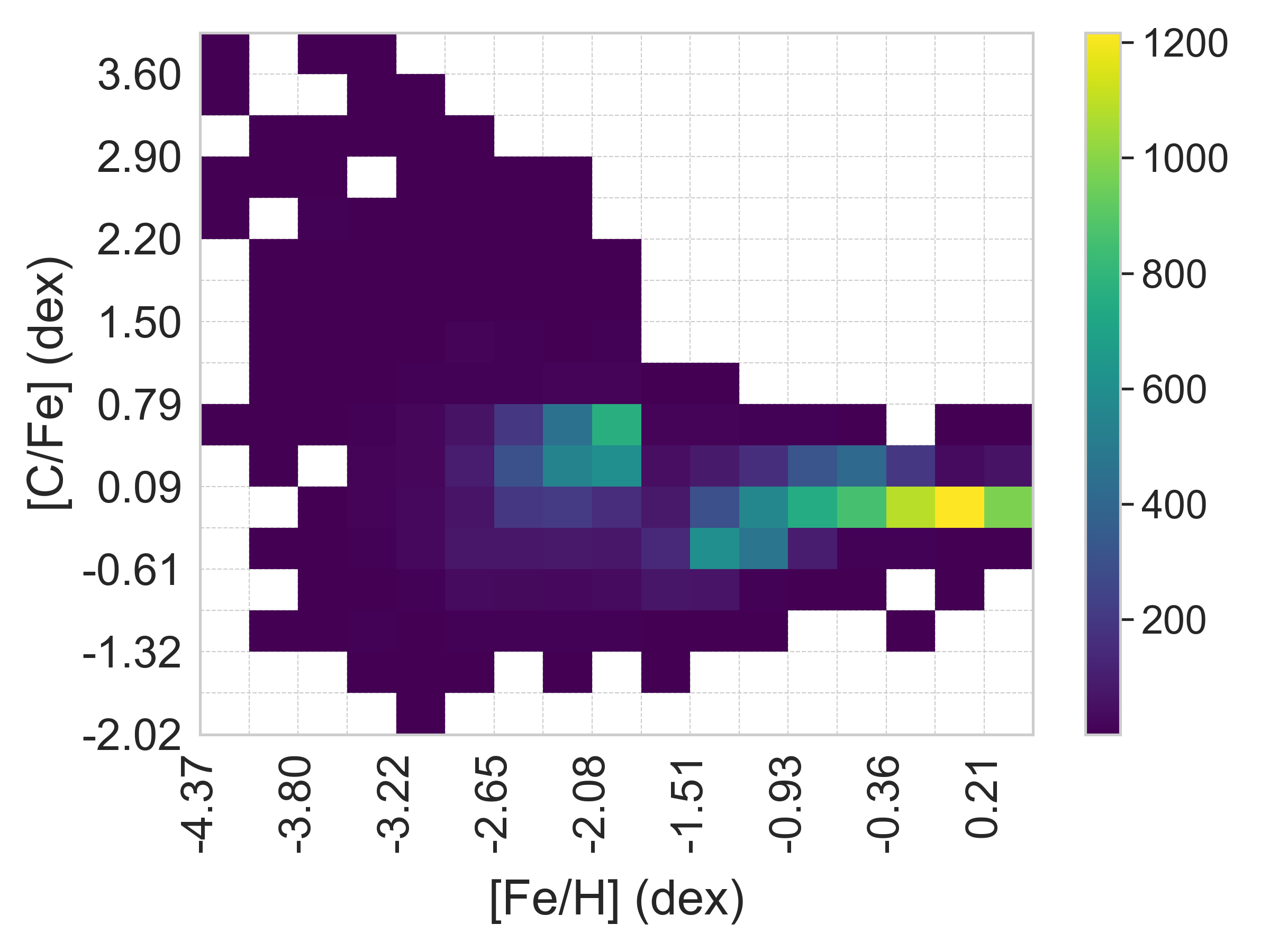}}
    \hfill
    \subfloat[{\scriptsize \texttt{[Fe/H]}-\texttt{[C/Fe]} error}\label{fig:feh_cfe_error}]{\includegraphics[width=0.48\linewidth]{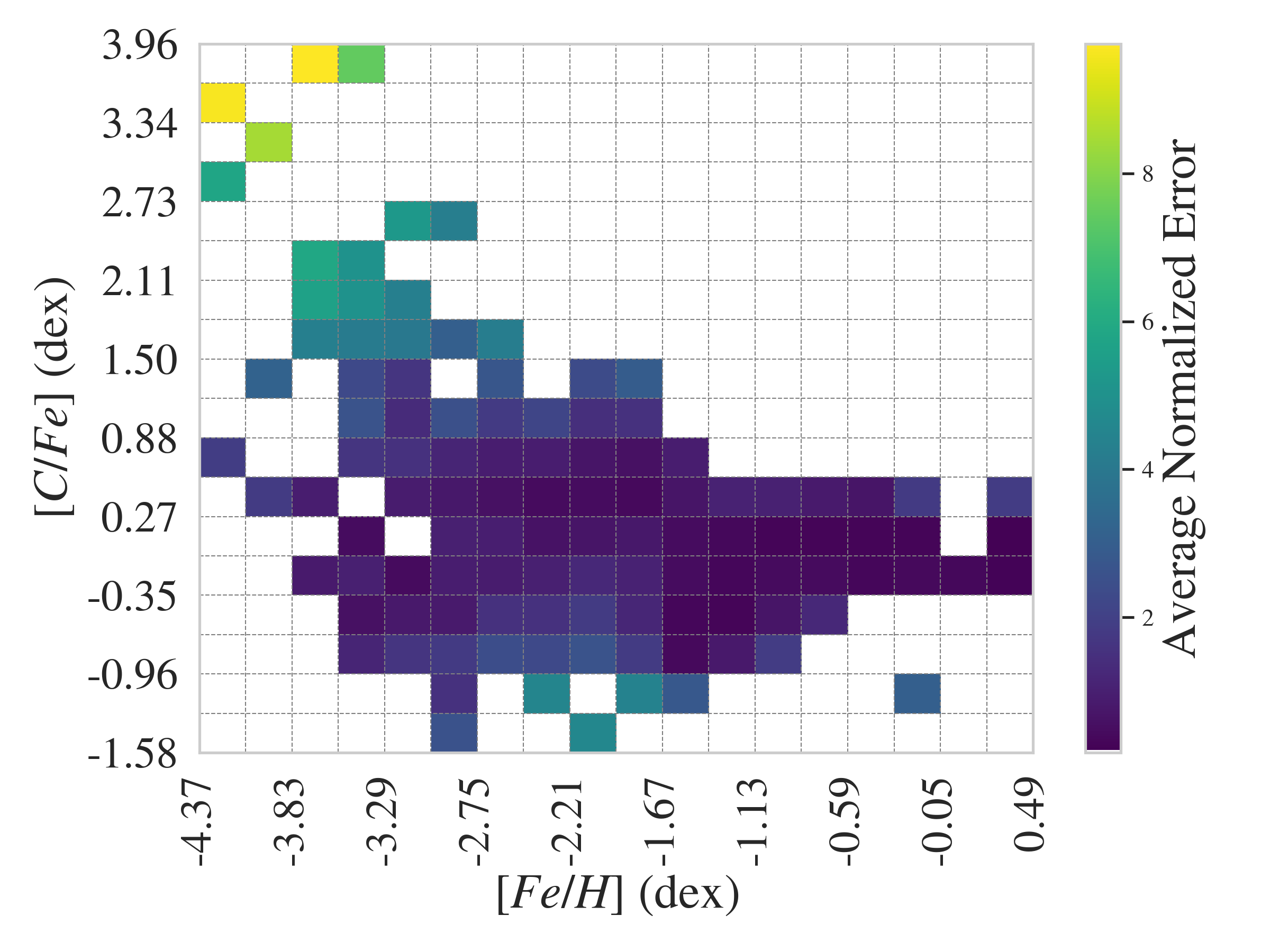}}
    
    \caption{Sample distribution and model estimation performance in [Fe/H]-[C/Fe] spaces. Panels (a) show the non-uniform sample distribution, while (b) demonstrate that parameter estimation errors increase as sample density decreases. Similar phenomena are observed in other parameter space and in the scenario of classification. The estimation errors are quantified as the average normalized absolute error sum.}
    \label{fig:distribution_density_prediction_accuracy}
\end{figure}

In regression studies, the WSLD dataset supports two research directions: single-output and multi-output regression. These two study directions must confront the challenges arising from low signal-to-noise ratios and uneven parameter distributions.
\begin{enumerate}
    \item Single-Parameter Regression: Focused on estimating a single parameter (e.g., $T_\texttt{eff}$, $\log~g$, [Fe/H], or [C/H]). Given that over 55\% of the samples in the dataset reside in low signal-to-noise ratio regions (Fig. \ref{fig:mae_snrg_distribution}(b)) and that sample distributions in key parameter spaces are highly uneven (Fig. \ref{fig:distribution_density_prediction_accuracy}(a)), models must be designed to extract features sensitive to weak signals under conditions of severe noise interference and sparse data in certain parameter regions, while enhancing robustness to imbalanced distributions to achieve more reliable parameter estimation.    
    \item Multi-Parameter Regression: Aimed at simultaneously estimating multiple parameters, this task requires designing multi-task learning models capable of capturing latent correlations among parameters. Given the highly uneven sample distribution in the multidimensional parameter space and the overall low signal-to-noise ratio, this task imposes heightened demands on the structural constraints of feature representations and their robustness to noise.
\end{enumerate}

For classification tasks, the WSLD dataset can be employed to model sample category differentiation; however, it faces challenges of severe class imbalance (with CEMP stars accounting for less than 2\%) and the concentration of low signal-to-noise ratio samples within minority classes (see Fig. \ref{fig:mae_snrg_distribution}(b)). These factors collectively limit the model’s ability to discern subtle inter-class differences. The scarcity of CEMP stars, coupled with their high astrophysical significance, necessitates models equipped with efficient feature extraction and selection mechanisms to enhance recognition of weak signals and rare classes, thereby mitigating performance degradation in sparse regions.

WSL represents a core challenge across both regression and classification tasks. However, the WSL exhibits distinct characteristics in each kind task. In regression tasks, weak features are often difficult to capture due to strong noise interference, leading to significantly degraded prediction performance in low signal-to-noise ratio regions, as illustrated in Fig. \ref{fig:mae_snrg_distribution}(a). In classification tasks, the scarcity of weak signals further limits the model’s ability to discriminate subtle inter-class differences. Although the high-dimensional feature structure (3,450 features) provides opportunities for uncovering latent patterns, it also imposes stricter requirements on feature selection and noise suppression. These challenges necessitate model designs that balance sensitivity in feature extraction with robustness to noise. Thereby, this dataset provides a solid foundation for WSL studies in complex scenarios.

\section{A benchmark method: Parallel Dual-View Fusion Network (PDVFN)}
\label{sec:methods}

This work constructs an astronomical spectral weak feature learning dataset named WSLD in section \ref{sec:DataSet}. Each sample in this dataset is represented as a vector whose components correspond to the observed energy/flux of a star at specific wavelength positions. The WSLD dataset holds significant value for WSL research, characterized by its low signal-to-noise ratio distribution skewness, dual imbalance in both parameter and class spaces, and high-dimensional feature structure (see analysis in Section \ref{sec:DataSet_statisticsAndAnalysis}). These characteristics provide a complex and realistic experimental environment for exploring regression, classification, and WSL tasks. These properties not only substantially increase the difficulty of model design but also present crucial opportunities for developing robust models capable of handling low SNR, sparse categories, and high-dimensional features. This effort contributes to advancing cutting-edge technologies that rely on weak signal modeling, such as astronomical spectrum analysis, medical image processing, and autonomous driving.

To support further research, this paper proposes a baseline learning method employing a dual-view approach. One view corresponds to the sample's original vector, while the other is a time-frequency image derived via short-time Fourier transform \cite{esmaeilpour2022multidiscriminator}. These views are hereafter referred to as vector representation and time-frequency representation, respectively.

\subsection{ Dual-View Representation and Overall Architecture }
\label{sec:methods:representation_framework}

The complementary nature of this dual-view representation is profoundly demonstrated through its synergistic mining of the intrinsic information structure within astronomical spectral signals. The vector representation preserves the original physical information of the spectrum, where its sequential structure directly corresponds to the physical relationship between wavelength and flux. It accurately retains the position, contour, and local morphology of spectral lines. The ACR module processes this sequence using convolutional, attention, and recurrent neural network mechanisms, focusing on capturing local spectral features—such as absorption and emission lines—and their sequential dependencies along the wavelength axis. This makes it particularly suitable for analyzing localized patterns with clear physical interpretations. In contrast, the time-frequency representation reveals the frequency-domain structural information of the spectral signal. By converting the one-dimensional spectrum into a two-dimensional time-frequency image via short-time Fourier transform, it translates localized mutations, periodic patterns, and non-stationary features from the wavelength domain into spatial patterns in the frequency domain. The PMTF module, through its multi-scale convolutional and state-space modeling, excels at capturing global frequency-domain distributions, energy spectrum trends, and implicit correlations across wavelengths. This allows it to identify distributed weak features that are difficult to detect in the original sequence. Together, these two views form a perfect complementarity between local and global, time-domain and frequency-domain, explicit physical features and implicit structural patterns. They jointly construct a more comprehensive and robust feature representation for astronomical spectra, effectively addressing the challenges of low signal-to-noise ratio and feature sparsity in the WSLD dataset.

The characteristics of the dataset—low signal-to-noise ratio, skewed distribution, and dual imbalance—make it difficult for traditional single-view feature learning methods to obtain stable and discriminative feature representations. Specifically, under low signal-to-noise conditions, meaningful signals are often drowned in noise and exhibit variable morphologies across different samples. Meanwhile, the dual imbalance in both class and parameter spaces demands that models possess strong robustness to sparse samples and rare patterns. Therefore, we propose a dual-view representation learning paradigm, fundamentally motivated by introducing two complementary feature perspectives with different inductive biases to construct a more redundant and robust feature representation space. This approach addresses the vulnerability of single feature streams in such complex data environments.

Accordingly, this work designs highly specialized feature extraction modules tailored for each view. For the vector view, the objective is to precisely capture subtle morphological variations at local wavelength positions (e.g., absorption lines, emission lines) and their long-range contextual dependencies in spectral sequences. This is achieved through the Attention Calibration Recurrent (ACR) module, which integrates convolutional, attention, and recurrent neural network mechanisms. For the time-frequency view, the aim is to uncover global characteristics—such as frequency-domain structures and energy distribution patterns—that are difficult to observe directly in the raw sequence. This is realized via the PMTF module, which incorporates multi-scale perception and state-space modeling capabilities. The motivation for fusing features from both modules is to enable the model to synergistically leverage local physical attributes and global structural patterns, forming a holistic perceptual capacity that ``sees both the trees and the forest." Through end-to-end training, the model learns to dynamically and adaptively extract and utilize information from the most beneficial perspectives based on input sample characteristics, ultimately enhancing predictive reliability under challenging conditions.

In summary, the overall architecture of the proposed PDVFN is structured as follows: the vector representation and time-frequency image representation of samples are processed by the Attention Calibration Recurrent (ACR) module and the Parallel Multi-scale Time-Frequency State Space (PMTF) module for feature extraction, respectively. The features output by these two modules are then concatenated into a unified vector and fed into a fully connected network to generate final classification predictions or parameter estimation results. Section \ref{sec:methods:ACR} details the ACR module, Section \ref{sec:methods:PMTF} elaborates on the PMTF module, and Section \ref{sec:loss_function} explains the loss functions used for model training. The ACR module achieves context-aware, attention-focused, and robust sequence representation extraction through local feature enhancement, sequential dependency modeling, and noise suppression. The PMTF module accomplishes scale invariance, long-term dependency modeling, and time-frequency structure identification via multi-scale structure capturing, frequency-domain feature extraction, and global pattern perception.

\subsection{Attention Calibration Recurrent Module (ACR)}
\label{sec:methods:ACR}

The ACR module achieves multi-stage attention calibration and sequential context modeling. It first utilizes two 1D convolutional layers to extract local positional-morphological features from spectral sequences, while employing CBAM's channel and spatial attention mechanisms to dynamically recalibrate feature weights—enhancing information-rich regions and suppressing noise. Subsequently, it further refines feature abstraction through secondary convolutional layers and CBAM reinforcement. Finally, a four-layer bidirectional GRU captures long-range contextual dependencies, combined with multi-head attention to focus on critical sequence segments. The processed features are aggregated through pooling and fully-connected layers to generate a highly condensed, task-oriented feature vector. This pipeline effectively addresses low signal-to-noise ratio and high-dimensional feature challenges, enabling end-to-end learning from raw spectral sequences to robust feature representations.

\subsection{Parallel Multi-scale Time-Frequency State Space Module (PMTF)}
\label{sec:methods:PMTF}

The PMTF module constructs a complete processing pipeline for multi-scale parallel fusion and time-frequency state space modeling. Initially, through parallel processing by the Multi-scale Feature Pyramid Fusion Unit (MFPF Unit) and the Reverse Convolution Block (RC Block) \cite{huang2025reverse}, complementary time-frequency features are extracted from spatial scales and channel dimensions respectively—the pyramid module captures multi-level time-frequency patterns from macroscopic to microscopic perspectives, while the reverse convolution block achieves feature refinement and recalibration through bottleneck structures and normalization operations. Subsequently, the outputs of these two modules are concatenated and fused along the channel dimension, forming a comprehensive feature representation with richer information.

The fused features are further processed by the Frequency-aware Feature Compression Unit (FFC Unit), which utilizes adaptive pooling and attention mechanisms to achieve frequency-selective feature enhancement and spatial regularization. Finally, four consecutive MambaVision modules \cite{hatamizadeh2025mamba} perform progressive sequence modeling on the compressed features, progressively establishing long-range time-frequency dependencies through state-space mechanisms to complete the full evolution path from local feature perception to global contextual understanding. The entire architecture combines the multi-scale spatial perception capabilities of CNNs with the dynamic sequence modeling advantages of state-space models, forming an end-to-end intelligent processing solution for time-frequency signals.

The Multi-scale Feature Pyramid Fusion (MFPF) Unit achieves the effect of multi-scale feature pyramid fusion. It extracts time-frequency features from three parallel processing paths at different resolution scales: the first path performs downsampling followed by upsampling to capture macroscopic contextual information; the second path extracts medium-detail features at the original scale; the third path performs upsampling followed by downsampling to capture microscopic fine structures. Finally, the three sets of features are concatenated along the channel dimension, forming a multi-scale feature representation spanning from coarse to fine granularity. This effectively enhances the model's joint perception capability for both long-term trends and transient features in time-frequency representations.

The Reverse Convolution (RC) Block achieves deep feature refinement and channel recalibration effects. It first applies LayerNorm for sequence normalization to stabilize training, then employs 1×1 convolutions for channel dimension expansion and reduction. Combined with 5$\times$5 reverse convolution \cite{huang2025reverse} to capture local spatial features, it forms a bottleneck structure. An intermediate LayerNorm layer further stabilizes feature distributions, while BatchNorm ensures output stability. The entire process maintains resolution unchanged while achieving intelligent channel compression and feature refinement, effectively enhancing the model's feature representation capability while controlling computational complexity.

\begin{figure}
    \centering
     \setlength{\belowcaptionskip}{-55mm} \includegraphics[width=0.8\linewidth]{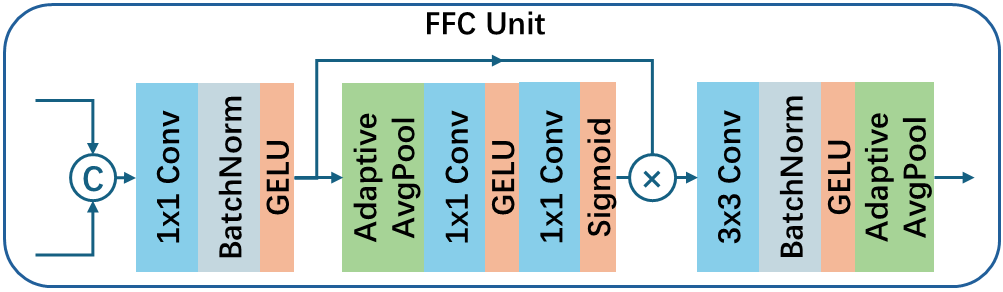}
    \caption{Framework of the Frequency-aware Feature Compression (FFC) Unit, a core component within the PMTF module of PDVFN. The core innovation is a frequency attention mechanism that generates frequency weights through adaptive average pooling along the temporal axis, followed by 1×1 convolutions with a sigmoid activation. These weights recalibrate the fused multi-scale and reverse convolution features, selectively emphasizing informative frequency components.    The compressed output is then resized via 3×3 convolution and adaptive pooling for subsequent state-space modeling.}   
    \label{fig:FFC_Unit_Framework}
\end{figure}

\subsection{Loss Function}
\label{sec:loss_function}

To enhance the WSL capability of the PDVFN under low signal-to-noise ratio scenarios, this work proposes two dedicated loss functions for parameter estimation and classification tasks respectively. The framework employs a robust loss function $L_{param}$ to stabilize regression training, while utilizing focal loss $L_{focal}$ to mitigate class imbalance. These two loss functions collectively strengthen the model's learning effectiveness on challenging samples.

\subsubsection{Loss function $L_{param}$ for Parameter Estimation}

For the parameter regression task, this paper employs the Negative Log-Likelihood (NLL) as the loss function to enhance model robustness in high-noise and weak-feature scenarios. The corresponding calculation formula is as follows:
\begin{equation}
\mathit{L_{param}} = \frac{\log \sigma^2_{\theta}(x)}{2} + \frac{(y - \mu_{\theta}(x))^2}{2\sigma^2_{\theta}(x)}
\end{equation}
where $x$ and $y$ denote the model input and corresponding reference label, respectively; $\mu_{\theta}(x)$ and $\sigma^2_{\theta}(x)$ represent the predicted mean and variance, respectively; and $\theta$ signifies the model’s trainable parameters.

This loss function is derived from a Gaussian likelihood model of the observed value $y$: $y \sim \mathcal{N}(\mu_{\theta}(x), \sigma^2_{\theta}(x))$. Within this probabilistic framework, the model is designed to simultaneously output both the location parameter ($\mu$) and dispersion parameter ($\sigma^2$) for each prediction. This dual-output architecture compels the model to introspect—requiring: It not only to provide point estimates but also to evaluate prediction confidence. For weak-signal samples with blurred features or significant noise interference, the model struggles to determine precise $\mu$ values due to insufficient discriminative information. Under such scenarios, predicting a larger $\sigma^2$ becomes probabilistically justified, reflecting the understanding that the ``true value may reside within a broader interval." From an optimization perspective, when the residual term $(y-\mu_{\theta}(x))^2$ is large—a typical characteristic of weak-signal scenarios—increasing the variance estimate $\sigma^2_{\theta}(x)$ effectively reduces the second term of the loss function, albeit at the cost of increasing the first term $\frac{\log \sigma^2_{\theta}(x)}{2}$. Through balancing these two competing terms, the model ultimately learns to assign higher uncertainty measures to low-quality predictions.

\subsubsection{Classification Loss function $L_{focal}$}

For the classification task, the presence of class imbalance and the ambiguous or noise-prone feature distributions of certain categories hinder the model's ability to learn discriminative features effectively. To address this issue, we employ Focal Loss \cite{mukhoti2020calibrating} as the loss function, which simultaneously mitigates data imbalance and enhances learning on weak-signal samples.

\section{EXPERIMENTAL RESULTS AND DISCUSSION}
\label{Sec:experiments}

To systematically evaluate the performance and effectiveness of the proposed model in weak-signal scenarios, this section conducts a comprehensive experimental investigations based on the newly constructed WSLD benchmark dataset. We design a series of experiments to analyze the model’s prediction accuracy, robustness, and generalization capability from multiple perspectives in both parameter estimation and classification scenarios.

\subsection{Evaluation Configurations}
\label{Sec:experiments_implementation_metrics}

All experiments were conducted on  PyTorch 2.3.0 and Python 3.12. The models were optimized with AdamW \cite{loshchilov2017decoupled}, an initial learning rate of 0.0005, and a decay factor of 0.1 every 10 epochs. Training was performed for 40 epochs.

For regression parameter estimation tasks, we employed the mean ($\mu$), standard deviation ($\sigma$), and mean absolute error (MAE) of the differences between predicted and ground-truth values as evaluation metrics. The mean $\mu$ indicates the overall bias or consistency between predictions and ground-truth, $\sigma$ measures prediction stability, and MAE quantifies the average magnitude of prediction difference across test samples.

For classification tasks, we used the area under the receiver operating characteristic curve (AUC), F1-Score, G-mean, and Matthews Correlation Coefficient (MCC) to assess performance\cite{haixiang2017learning}. AUC reflects the model’s discriminative ability, with values closer to 1 indicating superior performance; F1-Score balances precision and recall; G-mean evaluates combined recall for positive and negative classes; and MCC provides a balanced measure of classification performance based on the confusion matrix.

\subsection{Regressions for Parameter Estimation}
\label{Sec:experiments_reg}

To systematically evaluate the model’s regression capability, this section focuses on assessing its performance from two perspectives—single-parameter and multi-parameter regression. 

Table~\ref{tab:model_prediction_comparison_multiple_output} presents the evaluation results of the proposed method compared with StarNet~\cite{fabbro2018application,zhang2019938}, StarGRUNet~\cite{li2023estimating}, RRNet~\cite{xiong2022model}, and MSFL-IECW Net~\cite{Fang_2025} on the WSLD dataset for multi-output regression tasks. Overall, the proposed method achieves the best mean absolute error (MAE) across all parameters and obtains the best performance in terms of mean error ($\mu$) and error standard deviation ($\sigma$) for most parameters. Therefore, these experimental results demonstrate the effectiveness of the Parallel Dual-View Fusion Network (PDVFN) in weak-signal modeling. Specifically, for $T_{\text{eff}}$, our method attains an MAE of 100.5066, representing a reduction of approximately 13.8\% compared with the second-best method, MSFL-IECW Net (116.6274). For $\log~g$, our method achieves an MAE of 0.2456, reducing the error by about 8.5\% relative to MSFL-IECW Net (0.2684). For [Fe/H], the proposed method obtains an MAE of 0.1487, an improvement of around 21.0\% over RRNet (0.1882). For [C/H], our method reaches an MAE of 0.2007, corresponding to a reduction of approximately 14.9\% compared with RRNet (0.2358).

The fundamental reason behind this improvement lies in the Parallel Dual-View Fusion Network, which effectively integrates complementary information from vector representations and time–frequency image representations. Therefore, this scheme significantly enhances weak-signal feature expressiveness and noise suppression performance. Compared with single-view methods (e.g., RRNet, which relies solely on vector-based representations), the proposed method enables feature extraction from diverse perspectives and resulting in more comprehensive and robust modeling of complex inter-variable relationships by fusing the original vector representation with the two-dimensional time–frequency image representation obtained via STFT. In the scenarios characterized by high noise or weak feature expression, the dual-view fusion mechanism can substantially improve the model’s sensitivity to weak signals and its prediction accuracy by extracting discriminative information from complex backgrounds. This provides an efficient and stable solution for multi-output regression tasks.

Specifically, in the predictions of $T_{\text{eff}}$ and $\log~g$, the proposed Parallel Dual-View Fusion Network (PDVFN) substantially reduces prediction errors by integrating the ACR and PMTF modules. These experimental results demonstrate the superior feature-capturing capability of the proposed scheme compared with traditional methods. By leveraging complementary information from vector representations and time–frequency image representations, PDVFN strengthens the modeling of subtle mapping relationships between input features and structural variables. In contrast, traditional approaches such as StarNet, RRNet, and MSFL-IECW Net struggle to characterize complex weak-signal correlations in high-noise scenarios. The struggling results in limited predictive performance. For the [Fe/H] and [C/H] prediction tasks, the proposed method achieves performance improvements of 21.0\% and 14.9\% over the next-best method, RRNet. These experimental results highlight its robustness in handling low signal-to-noise ratio samples and noise-sensitive variables. These variables heavily depend on subtle local features in the input, which are easily affected by observational noise or background interference. The  observational noise or background interference result in larger prediction errors in StarNet, RRNet, and MSFL-IECW Net. Our method enhances local spectral feature detection, sequence dependency modeling, and noise suppression through the convolutional layers, CBAM attention mechanism, and bidirectional GRU in the ACR module. Meanwhile, the PMTF module extracts and refines multi-scale features from time–frequency images via its multi-scale feature pyramid fusion unit, deconvolution blocks, and frequency-aware feature compression unit, and further captures long-range dependencies using the state-space mechanism of the MambaVision module. Finally, the concatenation and integration of features from both views significantly strengthen the model’s capability in weak-signal detection and modeling.

In conclusion, our method’s superior performance in multi-variable regression tasks validates the effectiveness of the proposed parallel dual-view fusion network. These experimental results not only establish a new benchmark for multi-output regression on the WSLD dataset but also provide new insights for future regression studies on weak-signal data.

\begin{table*}[!t]
    \centering
    \caption{Model Comparison on Parameter Prediction Capabilities in Multi-Parameter Regression}
    \label{tab:model_prediction_comparison_multiple_output}
    \resizebox{\textwidth}{!}{
        \begin{tabular}{@{}lcccccccccccc@{}}
            \toprule
                & \multicolumn{3}{c}{$T_{\text{eff}}$(K)} & \multicolumn{3}{c}{$\log~g$ (dex) } & \multicolumn{3}{c}{[Fe/H] (dex)} & \multicolumn{3}{c}{[C/H] (dex)} \\
                \cmidrule(lr){2-4} \cmidrule(lr){5-7} \cmidrule(lr){8-10}\cmidrule(lr){11-13}
                {Model Error} & {$\mu$} & {$\sigma$} & {MAE} & {$\mu$} & {$\sigma$} & {MAE} & {$\mu$} & {$\sigma$} & {MAE} & {$\mu$} & {$\sigma$} & {MAE} \\
            \midrule
                StarNet\cite{fabbro2018application,zhang2019938}{} & 29.7040 & 224.7114 & 164.1658 & 0.0998 & 0.4985 & 0.3713 & 0.0444 & 0.3726 & 0.2589 & 0.0968 & 0.4217 & 0.3219 \\
                StarGRUNet\cite{li2023estimating}      & -14.0612 & 207.2982 & 128.1916 & -0.0305 & 0.4770 & 0.3089 & -0.0152 & 0.3255 &  0.1906 & 0.0100 & 0.3962 & 0.2476 \\
                RRNet\cite{xiong2022model}          & \textbf{5.6162} & 192.7556 & 123.3502 & 0.0117 & 0.4424 & 0.3052 & -0.0090 & 0.3126 & 0.1882 & \textbf{0.0023} & 0.3711 & 0.2358 \\
                MSFL-IECW Net\cite{Fang_2025}          & 32.2358 & 180.2494 & 116.6274 & 0.0288 & 0.3996 & 0.2684 & 0.0554 & 0.2918 & 0.1914 & 0.0755 & 0.3471 & 0.2387 \\      
                Our           & -13.1643 & \textbf{171.8149} & \textbf{100.5066} & \textbf{-0.0088} & \textbf{0.3969} & \textbf{0.2456} & \textbf{0.0083} & \textbf{0.2864} & \textbf{0.1487} & 0.0048 & \textbf{0.3452} & \textbf{0.2007} \\                        
            \bottomrule
        \end{tabular}
    }
\end{table*}

We also experimentally compared the proposed method with StarNet~\cite{fabbro2018application,zhang2019938}, StarGRUNet~\cite{li2023estimating}, RRNet~\cite{xiong2022model}, and MSFL-IECW Net~\cite{Fang_2025} in the scenario on individual parameter estimation. It is shown that the proposed method achieves the best overall performance across all three evaluation metrics—mean error ($\mu$), standard deviation of error ($\sigma$), and mean absolute error (MAE)—for every predicted parameter. 

\subsection{Comparison on Classification}
\label{Sec:experiments_cls}

In this section, we conducted a detailed comparison of our Parallel Dual-View Fusion Network against a series of typical approaches in the scenerio of classification. For example, XGBoost \cite{2023MNRASLucey}, Convolutional Neural Network (CNN) \cite{Xie_2021}, and Artificial Neural Network (ANN) \cite{ardern2025predicting}. The results, presented in Table \ref{tab:model_Cls_comparison}, demonstrate that our method significantly outperforms others across all metrics: AUC of 0.9610, F1-Score of 0.7391, G-mean of 0.8627, and MCC of 0.8233. Compared with XGBoost, CNN, and ANN, our method improves AUC by 33.0\%, 25.8\%, and 17.1\%; F1-Score by 46.4\%, 21.9\%, and 7.6\%; G-mean by 28.1\%, 17.9\%, and 6.8\%; and MCC by 86.2\%, 6.9\%, and 2.7\%, respectively. These results highlight our method’s superior classification performance, particularly in weak signal detection tasks with high noise and class imbalance.

The core advantage of our method lies in the dual-view representation strategy embedded within the Parallel Dual-View Fusion Network. The dual-view representation makes it possible to fully leverage the complementary information from the original one-dimensional vector representation together with the two-dimensional time–frequency image representation derived from STFT from multiple perspectives, overcome the limitations of single-view representation. In contrast, traditional methods typically rely on a single-view input and therefore struggle to capture dynamic temporal information or local variation patterns. These limitations result in constrained performance when facing complex background noise or subtle feature structures.

A closer examination reveals that XGBoost performs substantially worse than deep learning models. As a tree-based algorithm, it offers interpretability and moderate capability in simple tasks, but its expressiveness is insufficient for the high-dimensional, nonlinear data in WSLD. It cannot model temporal dependencies or fine-grained local patterns, resulting in pronounced weaknesses under low SNR or imbalanced conditions with complex background noise.

CNN and ANN outperform XGBoost in classification. This result indicates that deep neural networks are capable of extracting useful features through end-to-end learning. However, both approaches still exhibit notable limitations. First, they rely solely on one-dimensional views and thus fail to exploit implicit feature representations in the frequency or spatial domains. This failure makes it difficult for the model to accurately extract weak signal features under complex background interference. Second, the relatively simple architectures of CNN and ANN constrain their ability to classify weak signals in noisy environments. 

In contrast, the proposed PDVFN thoroughly considers the characteristics of weak signals, leverages the advantages of different data views through a parallel dual-view fusion network. Specifically, the ACR module effectively captures sequential dependencies and local detailed patterns from the raw sequences via its attention mechanism and recurrent neural network architecture. Meanwhile, the PMTF module focuses on time-frequency image: it first extracts and enhances weak yet discriminative frequency-domain features through a multi-scale feature pyramid fusion unit and deconvolutional blocks, and further employs its internal MambaVision module to perform efficient global context modeling on the feature sequences via a selective state space mechanism. Finally, features from the dual views are integrated through concatenation, enabling the model to utilize complementary information from both temporal and frequency domains, thereby achieving a holistic understanding of the signal. Experimental results validate the effectiveness of the proposed approach and establish a new benchmark for future classification research on low signal-to-noise ratio sequential data.

\begin{table}[!t]
    \centering
    \caption{Comparison based on Classification Results}
    \label{tab:model_Cls_comparison}
    \resizebox{\linewidth}{!}{
        \begin{tabular}{lcccc}
            \toprule
            Model & AUC & F1 Score & G-mean & MCC \\ 
            \midrule
            XGBoost\cite{2023MNRASLucey} & 0.7226 & 0.5049 & 0.6736 & 0.4422 \\
            CNN\cite{Xie_2021}            & 0.7640 & 0.6064 & 0.7317 & 0.7700 \\
            ANN\cite{ardern2025predicting} & 0.8207 & 0.6870 & 0.8075 & 0.8018 \\
            Our                             & \textbf{0.9610} & \textbf{0.7391} & \textbf{0.8627} & \textbf{0.8233} \\            
            \bottomrule
        \end{tabular}
    }

    \vspace{1ex}
    \noindent
    \begin{minipage}[t]{\dimexpr\linewidth-2\tabcolsep\relax}
        \footnotesize\textit{Note: The best value for each metric is indicated in bold.}
    \end{minipage} 
\end{table}

\section{CONCLUSION}
\label{Sec:conclusion}

This study systematically investigates the key challenges in WSL—low signal-to-noise ratio and extreme class imbalance—with particular focus on accurate identification and prediction of ``critical minority samples." To address the lack of specialized evaluation benchmarks in this field, we developed WSLD, the first dedicated dataset for WSL. Comprising 13,158 high-dimensional samples, WSLD exhibits some typical challenges including dominant low SNR and dual parameter-class imbalance. Therefore, this dataset establishes a unified benchmark for evaluating model generalization on edge cases.

To support subsequent research, this paper proposes a dual-view representation scheme and a Parallel Dual-View Fusion Network (PDVFN) as a baseline solution, specifically designed for the characteristics of weak-signal datasets including low signal-to-noise ratio, distribution skewness, and dual imbalance. The approach systematically constructs dual views from the original vector and its time-frequency transformed image, employs ACR and PMTF modules to concurrently extract local sequential features and global frequency-domain structures. The complementary nature of the dual views manifests in their synergistic exploration of time-frequency domain information in spectral signals. By leveraging multi-source information fusion, PDVFN effectively enhances the model's representational capacity under low signal-to-noise ratios and imbalanced data conditions, providing an innovative solution for weak-signal learning tasks.

Experimental results demonstrate that the proposed PDVFN method significantly outperforms existing state-of-the-art benchmarks in both parameter regression and classification tasks. The PDVFN exhibits exceptional performance particularly on low signal-to-noise ratio samples and minority class instances.

\bibliographystyle{IEEEtran}
\bibliography{cites}

\vfill

\end{document}